\documentclass{article}

     \PassOptionsToPackage{numbers, compress}{natbib}
     \usepackage[preprint]{neurips_2019}

\usepackage[utf8]{inputenc} % allow utf-8 input
\usepackage[T1]{fontenc}    % use 8-bit T1 fonts
\usepackage{hyperref}       % hyperlinks
\usepackage{url}            % simple URL typesetting
\usepackage{booktabs}       % professional-quality tables
\usepackage{amsfonts}       % blackboard math symbols
\usepackage{nicefrac}       % compact symbols for 1/2, etc.
\usepackage{microtype}      % microtypography
\usepackage{amsmath}
\usepackage{graphicx}       % added by Marshall 2021-04-16
\usepackage{float}          % added by Marshall 2021-04-16
\usepackage{wrapfig}        % added by Ryan 2021-04-18

\usepackage{natbib}         % added by Marshall 2021-04-16
\title{When to Fold'em: \\How to answer Unanswerable questions}

\author{%
  Marshall Ho, Zhipeng Zhou, Judith He\\
  Department of Computer Science\\
  University of Toronto\\
  \texttt{marshall.ho@utoronto.ca,\{zhipeng.zhou,judith.he\}@mail.utoronto.ca} \\
}

\begin{document}

\maketitle

\begin{abstract}

    We present 3 different question-answering models trained on the SQuAD2.0 dataset [\citenum{noauthor_stanford_nodate}]-- BIDAF [\citenum{seo_bidirectional_2018}], DocumentQA [\citenum{clark_simple_2017}] and ALBERT Retro-Reader [\citenum{zhang_retrospective_2020}]-- demonstrating the improvement of language models in the past 3 years.   Through our research in fine-tuning pre-trained models for question-answering, we developed a novel approach capable of achieving a 2\% point improvement in SQuAD2.0 F1 in reduced training time.  Our method of re-initializing select layers of a parameter-shared language model is simple yet empirically powerful.

\end{abstract}

\section{Introduction}
    The field of natural language processing has undergone major improvements in recent years.  The replacement of recurrent architectures by attention-based models has allowed NLP tasks such as question-answering to approach human level performance.  In order to push the limits further, the SQuAD2.0 dataset was created in 2018 with 50,000 additional unanswerable questions, addressing a major weakness of the original version of the dataset (SQuAD1.1 [\citenum{rajpurkar_squad_2016}]).  Together with the advent of pure attention-based models pre-trained on massive corpuses (BERT [\citenum{devlin_bert_2019}], RoBERTa [\citenum{liu_roberta_2019}], XLNet [\citenum{yang_xlnet_2020}]), the state of the art language models (ALBERT [\citenum{lan_albert_2020}], ELECTRA [\citenum{clark_electra_2020}], SpanBERT [\citenum{joshi_spanbert_2020}] ) regularly exceed human level performance in question-answering benchmarks.

%---------------------------------------------------------------------------------
\section{Related Work}
    The original version of SQuAD [\citenum{rajpurkar_squad_2016}] was published in 2016.  There are also a variety of benchmarks related to question-answering -- NewsQA [\citenum{trischler_newsqa_2017}], SearchQA [\citenum{dunn_searchqa_2017}], TriviaQA [\citenum{joshi_triviaqa_2017}], HotpotQA [\citenum{yang_hotpotqa_2018}], and Natural Questions [\citenum{kwiatkowski_natural_2019}].  Nonetheless, SQuAD is one of the most commonly used question-answering benchmarks.  For example, the seminal BERT paper [\citenum{devlin_bert_2019}] relied only on SQuAD to demonstrate the model’s question-answering ability. With respect to natural language models, recent work which did particularly well with the SQuAD 2.0 dataset includes ALBERT [\citenum{lan_albert_2020}], Retrospective Reader [\citenum{zhang_retrospective_2020}] -- both featured in this report -- as well as ELECTRA [\citenum{clark_electra_2020}] and SpanBERT [\citenum{joshi_spanbert_2020}].

%---------------------------------------------------------------------------------
\section{Models and Methods}
  % Introduce the structure of all models, without data and experiment included

\subsection{BIDAF}
    In 2016, Bi-Directional Attention Flow (BIDAF) [\citenum{seo_bidirectional_2018}] was introduced. Based on the function of different layers, we divide 6 layers into 3 parts. There are 3 layers in the first part, \verb|Character Embedding|, \verb|Word Embeddin|, and \verb|Contextual Embedding Layer|. They embed the words from both query and context to two attention flow,  $\mathbf{H}\in\mathcal{R}^{2d\times T}$ for context and $\mathbf{U}\in\mathcal{R}^{2d\times J}$ for query, where $T$ and $J$ represents the number of words in the input context and query.
    
    The second part is the core of BIDAF, including \verb|Attention Flow Layer| and \verb|Modeling Layer|. Compared with traditional attention mechanisms, BIDAF uses bi-directional attentional flow, which not only attends query on the context (Q2C) but also context to query (C2Q). In this case, the attention information from both sides complements each other. The model first calculates the similarity between each context word and query word $\mathbf{S}_{tj}$,
    % \[
    %     \mathbf{S}_{tj}=\alpha(\mathbf{H}_{:t},\mathbf{U}_{:j})\in\mathcal{R}
    % \]
    and feeds it to \verb|softmax| layer to get new attention flows $\hat{\mathbf{U}}$ and $\hat{\mathbf{H}}$. Two new attention flows are combined by a multi-layer perceptron (MLP), and passed into two layers of bi-directional LSTM [\citenum{hochreiter_long_1997}]. In the end, it outputs a matrix $\mathbf{M}\in\mathcal{R}^{2d\times T}$, where each column vector contains contextual information between context and query.
    
    The last part contains only an \verb|output layer|, which produces the probability distribution of start index $\mathbf{p}^1$ and end index $\mathbf{p}^2$ over the paragraph ($\mathbf{p}^i=softmax(\mathbf{w}^\top_{(\mathbf{p}^i)}[\mathbf{G;M}])$). $^\ast$Please refer to the Appendix for details about the BIDAF model.
    
\subsection{DocumentQA}
For the Document QA model [\citenum{clark_simple_2017}], there are 6 layers: input layer, embedding layer, pre-processing layer, attention layer, self-attention layer and prediction layer. The detailed explanation of the first three layers is listed in the appendix. We focus on the core parts of the model. In the attention layer, it builds a question-aware context representation by a bi-directional attention mechanism from the BIDAF model. In that layer, we learn the weights $w_1$, $w_2$, $w_3$ in the attention equation listed below. The weights imply the inner connections between the question and the context.

Attention between context word i and question word j: 
$a_{ij} = w_1\cdot h_i+w_2\cdot q_j+ w_3\cdot(h_i\odot q_j$)\\
($h_i$: vector for the context word $i$, $q_j$: vector for the question word $j$, $n_q$, $n_c$: lengths for the question and context respectively)

Then we output vectors representing the relationship between context and questions to the self attention layer. The output vectors are listed in the appendix. The output vectors can help the self-attention layer to understand the inner relationship between the words in the passage such as semantics and logics. Then we feed the output of attention layer, implying the relationship between question and passage, and the output of self-attention layer, implying the inner relationship of the passage itself into the prediction layer. The prediction layer then predicts the most likely answer start token and answer end token based on the scores each token receives.\\

\subsection{ALBERT and the Retro-Reader}
\subsubsection{ALBERT}
The goal of ALBERT (A Lite BERT) [\citenum{alberti_bert_2019}] is to reduce the memory requirement of the ground-breaking language model BERT [\citenum{devlin_bert_2019}], while providing a similar level of performance.  ALBERT uses two methods to reduce the number of parameters -- parameter sharing and factorized embedding.  Please refer to the Appendix for an in-depth analysis of these two methods.

\subsubsection{Retro Reader}
Near the top of the SQuAD2.0 leaderboard [\citenum{rajpurkar_know_2018}] is Shanghai Jiao Tong University’s Retro-Reader [\citenum{zhang_retrospective_2020}].  We have re-implemented their non-ensemble ALBERT model, as shown in Fig \ref{fig:albert03RetroReader} in the Appendix.

\subsubsection{Novel fine-tuning of a shared-parameter model}
ALBERT is pre-trained with the masked language modeling (MLM) objective similar to BERT.  A certain percentage of tokens are randomly masked and the model’s job is to predict the masked tokens correctly.  This type of token-prediction training, however, is not ideal for the task of span prediction [\citenum{joshi_spanbert_2020}], such as SQuAD2.0’s Question and Answering.

\begin{figure}[h]
    \centering
    \begin{minipage}{0.45\textwidth}
        \centering
        \includegraphics[width=0.8\textwidth]{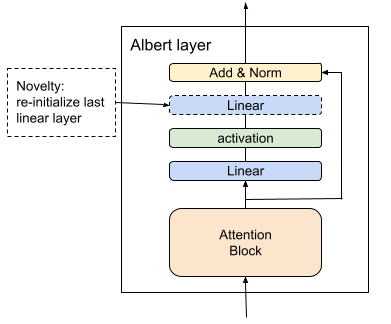}
        \caption{Re-initialize the last linear layer of ALBERT to perform better at SQuAD2.0.}
        \label{fig:albert04zeroWeights}
    \end{minipage}\hfill
    \begin{minipage}{0.45\textwidth}
        \centering
        \includegraphics[width=0.9\textwidth]{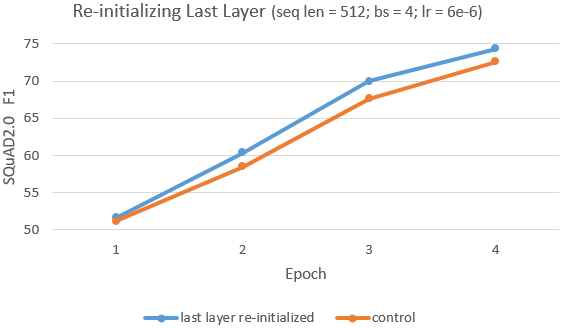}
        \caption{Novel fine-tuning by specific re-initializing of a parameter-shared network}
        \label{fig:albert14novel}
    \end{minipage}
\end{figure}

The last layers of ALBERT are likely overfitted to the MLM task.  We hypothesize that, by re-initializing the weights of the last layer of the ALBERT model, we can save time “un-learning” the MLM task.  However, because the weights for all layers are shared in our base ALBERT model, there is no clear way of simply re-initializing the “last” transformer layer.  We studied the structure of ALBERT in details [\citenum{lan_albert_2020}] and proceeded to re-initialize the last linear layer of the ALBERT transformer.  Fig \ref{fig:albert04zeroWeights} depicts our novel approach.

%---------------------------------------------------------------------------------
\section{Experiments}
  % All experiments, comparison, and analyse will be included here

\subsection{BIDAF in SQuAD1.1 \& SQuAD2.0}
\begin{figure}[h]
      \includegraphics[scale=0.35]{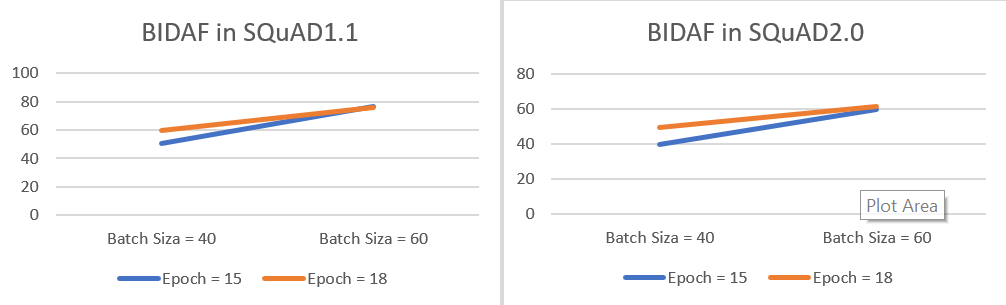}
      \centering
      \caption{BIDAF: tuning batch size and epoch in different datasets}
      \label{fig:bidafexper}
    \end{figure}

    To test the difference between two datasets and the influence of datasets, we performed hyperparameter tuning using BIDAF model in both SQuAD1.1 and SQuAD2.0. We have based our implementation of the BIDAF on the code released at https://github.com/allenai/bi-att-flow and https://github.com/zhuzhicai/SQuAD2.0-Baseline-Test-with-BiDAF-No-Answer, and our modifications are mainly the change of reading and writing pass and the hyper-parameters. Since it takes a long time to train the model (8-12 hours in 2080Ti GPU), we only trained the model in each datasets four times (Fig \ref{fig:bidafexper}).
    
    Here we can find, obviously, compared with the performance of BIDAF in SQuAD1.1, BIDAF performed terribly in SQuAD2.0. The reason why this happened may be because there are over 50,000 unanswerable questions more included in SQuAD2.0. However, there are only answerable questions (answer can always be found in context) in SQuAD1.1. According to the BIDAF model we described above, what we get eventually from BIDAF is the probability distribution of the start index and the end index in context, which means the model will always find an answer from the context. Therefore, in a dataset which includes unanswerable questions, the BIDAF model cannot perform well. 
    
    $^\ast$Please refer to the Appendix for the detail about the BIDAF experiment.

\subsection{DocQA}

 Document-level inputs are built by concatenating all the paragraphs from the same articles into 1 document in Squad1.1 [\citenum{clark_simple_2017}].
Since training a model with entire documents as input can be extremely computationally expensive, Docqa was built to process paragraph-level inputs sampling from the entire documents. However, we need a method that can produce accurate per-paragraph confidence scores during sampling paragraphs including the paragraphs without answers. The pros and cons for each confidence method are listed in the appendix.

In Table \ref{table:1}(hyper-parameters in Appendix), we find that our paragraph-level model performs much better than the BIDAF model which only gets 76.2 (best f1 score) on Squad1.1. We compared the structures between 2 models and found that adding an extra self-attention layer may help the model to better understand the logic inside the contexts. With more knowledge of the context itself, Docqa model can predict the correct answer spans given the query-aware context representation from the attention layer better. In addition to accuracy increase, Docqa requires much less training time than BIDAF because of our paragraph-level design. Furthermore, the confidence methods can handle the multi-paragraph sampling pretty well, especially for the shared normalization method.

\begin{table}[h!]
    \centering
     \begin{tabular}{||c c c c||} 
     \hline
     Model & Exact Match (EM) & F1 & Training Time \\
     \hline\hline
     None (paragraph-level) & 71.5137 & 80.4782 & 5.6 hrs \\ 
     Original Confidence & 70.9649 & 80.0798 & 7.4 hrs \\
     Shared Normalization & 71.4380 & 80.6665 & 9.7 hrs \\
     Merge & 70.6338 & 79.8808 & 8.5 hrs \\
     \hline\hline
    \end{tabular}
    \caption{Table to DocQA performance}
    \label{table:1}
\end{table}
\begin{wrapfigure}{r}{0.35\textwidth} %this figure will be at the right
    \includegraphics[width=0.35\textwidth]{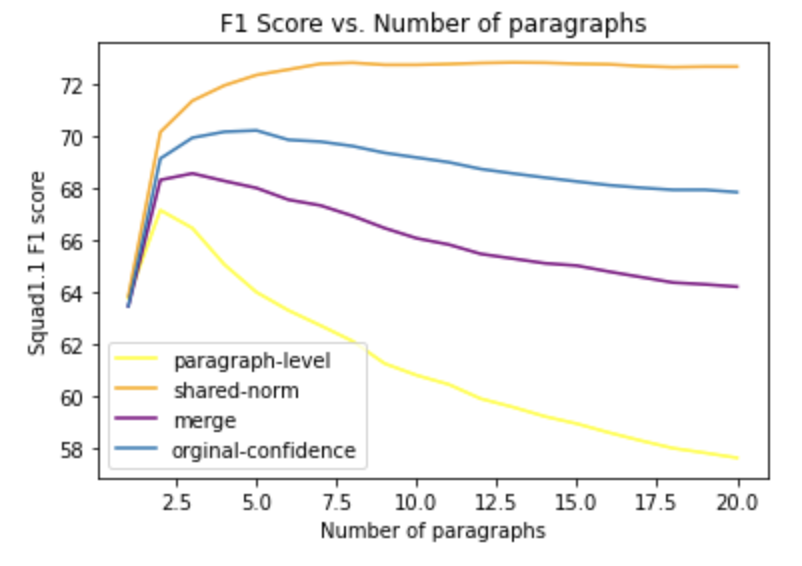}
      \centering
      \caption{Models of Docqa}
      \label{fig:docqamodels}
\end{wrapfigure}

Figure \ref{fig:docqamodels} shows the document-level performance on Squad1.1 with different modes. We find that the paragraph-level model perform very poorly once the number of paragraphs is increased. This is because it has only been trained on answer-containing paragraphs and it does not have confidence methods to compare which paragraph may contain the relevant information to the question. However, the shared normalization method still performed very well, and we believe this method can provide rather accurate confidence scores when sampling the paragraphs.
\subsubsection{Novel idea:}
In the prediction layer, the answer end token is predicted based on the hidden state of the answer start token. We think the prediction of the start token is effected by the end token in some way, so we add the hidden state of the end token to the prediction of the start token, as shown in the graph in Appendix. However, we find that there exists a "cycle" when doing back-propagation, causing error in the gradient updates.

\subsection{Retro-Reader with ALBERT}

Please refer to the Appendix for our analysis of the tuning of several hyperparameters for this model.

\subsubsection{Novel fine-tuning method: Specific re-initializing of a parameter-shared network}
We designed this experiment to examine whether re-initializing the layer closest to the masked token prediction head would help with fine-tuning the model for SQuAD2.0.  Fig \ref{fig:albert14novel} presents the comparison between our novel method and the control configuration.  Our method consistently provided an advantage of about 2\% points from epochs 2 to 4.  This confirms our hypothesis that our specific method of re-initializing a shared-parameter model can help “un-learn” the masked-token-prediction pre-training task, and speed up fine-tuning.
%---------------------------------------------------------------------------------
\section{Conclusion}
Our research covered the development of NLP in the past 3 years, when the attention mechanism largely displaced recurrent architectures, using the SQuAD2.0 dataset [\citenum{noauthor_stanford_nodate}] as the centerpiece of our discussion.  Through our experiments with BIDAF [\citenum{seo_bidirectional_2018}], DOCQA [\citenum{clark_simple_2017}] and ALBERT Retro-Reader [\citenum{zhang_retrospective_2020}], we examined the architectural differences and the advantages of the using state-of-the-art pretrained models and then fine-tuning for the question-answering task.  Our major contribution is our specific method of re-initializing weights in a parameter-shared network.  Future work can include applying our re-initialization method to Electra  [\citenum{clark_electra_2020}].
%---------------------------------------------------------------------------------

\newpage
\section{Appendix}

\subsection{BIDAF}
\subsubsection{Model}

\begin{wrapfigure}{r}{0.6\textwidth} %this figure will be at the right
    \includegraphics[width=0.6\textwidth]{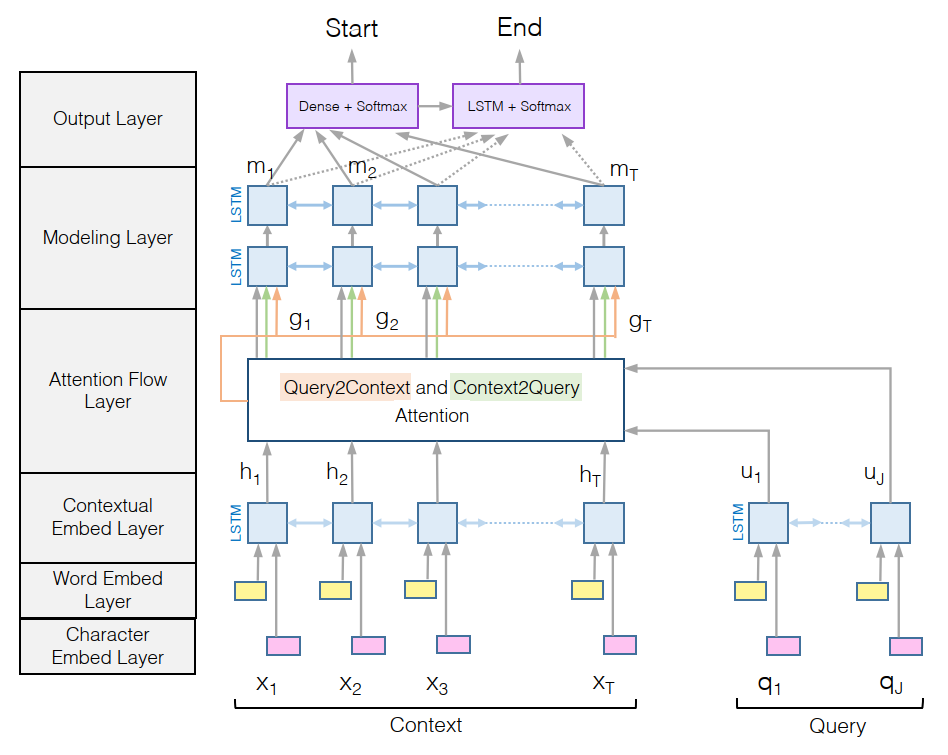}
      \centering
      \caption{BIDAF structure}
      \label{fig:bidstruc}
\end{wrapfigure}

    As we mentioned before, we divide the BIDAF model [\citenum{seo_bidirectional_2018}] into 3 part. In the first "embedding" part, there are \verb|Character Embedding|, \verb|Word Embeddin|, and \verb|Contextual Embedding| three layers. The words in both context and query are embedded in character-level and word-level respectively. In \verb|Character Embedding Layer|, the characters in each word are embedding into vector by Convolutional Neural Networks (CNN) [\citenum{kim_convolutional_2014}]. In \verb|Word Embedding Layer|, the BIDAF model uses pre-trained word vectors, GLoVe [\citenum{pennington_glove_2014}], which was trained based on Wikipedia data, to obtain the word embedding. And eventually, two bi-direction Long Short-Term Memory Networks (LSTM) [\citenum{hochreiter_long_1997}] was used in \verb|Contextual Embedding Layer| to get the temporal interaction between words in context and query respectively, which eventually output two attention flow, $\mathbf{H}\in\mathcal{R}^{2d\times T}$ for context and $\mathbf{U}\in\mathcal{R}^{2d\times J}$ for query, where $T$ and $J$ represents the number of words in the input context and query.
    
    The second part of the BIDAF model is the core part. The main approach to realize the function is, firstly, calculate the similarity between each word in context and query,
    \begin{align*}
        \mathbf{S}_{tj}&=\alpha(\mathbf{H}_{:t}, \mathbf{U}_{:j})\\
        &=\mathbf{w}^\top_{\mathbf{S}}[\mathbf{H};\mathbf{U};\mathbf{H}\odot \mathbf{U}]
    \end{align*}
    where $[;]$ is vector concatenation across row. And then use the similarity to calculate the C2Q attention and the Q2C attention,
    \begin{align*}
        &\text{C2Q:}\ \tilde{\mathbf{U}}=\sum_j\text{softmax}(\mathbf{S})\mathbf{U}\\
        &\text{Q2C:}\ \tilde{\mathbf{H}}=\sum_j\text{softmax}(\text{max}_{col}\mathbf{S})\mathbf{H}
    \end{align*}
    Then two new attentions are combined by a MLP $\beta$ to yield $\mathbf{G}$ in \verb|Attention Flow Layer|,
    \begin{align*}
        \mathbf{G}&=\beta(\mathbf{H};,\tilde{\mathbf{U}},\tilde{\mathbf{H}})\\
        &=[\mathbf{H};\tilde{\mathbf{U}};\mathbf{H}\odot\mathbf{\tilde{\mathbf{U}}};\mathbf{H}\odot\mathbf{\tilde{\mathbf{H}}}]\in\mathcal{R}^{8d\times T}
    \end{align*}
    The output $G$ will then pass into a two layers bi-directional LSTM to find the interactions between content and query in \verb|Modeling Layer|. And we will obtain a matrix $M$ in the end, where each column vector contains contextual information between context and query.
    
    Eventually, the output layer produces the probability distribution of start index and end index over the paragraph.
    
    The detailed structure of BIDAF is in Figure \ref{fig:bidstruc}, which is from Minjoon et al [\citenum{seo_bidirectional_2018}].
    \newpage
\subsubsection{Hyper-parameters tuning}
    We have based our implementation of the BIDAF on the code released at https://github.com/allenai/bi-att-flow and https://github.com/zhuzhicai/SQuAD2.0-Baseline-Test-with-BiDAF-No-Answer, and our modifications are mainly the change of reading and writing pass and the hyper-parameters. Since the model takes around 8-12 hours to train the model in full SQuAD dataset, we only trained the model in each datasets four times to tune the hyper-parameter (Table \ref{table:2}).
    
    \begin{table}[h!]
     \begin{tabular}{||c c c c||} 
     \hline
     SQuAD1.1 & Batch Size & Epoch & F1 \\ [0.5ex] 
     \hline\hline
     1 & 40 & 15 & 50.2 \\ 
     \hline
     2 & 40 & 18 & 59.6 \\
     \hline
     3 & 60 & 15 & 76.3 \\
     \hline
     4 & 60 & 18 & 76.2 \\
     \hline
    \end{tabular}\ 
    \begin{tabular}{||c c c c||} 
     \hline
     SQuAD2.0 & Batch Size & Epoch & F1 \\ [0.5ex] 
     \hline\hline
     1 & 40 & 15 & 39.8 \\ 
     \hline
     2 & 40 & 18 & 49.6 \\
     \hline
     3 & 60 & 15 & 59.4 \\
     \hline
     4 & 60 & 18 & 61.2 \\
     \hline
    \end{tabular}
    \caption{Table to BIDAF on SQuAD1.1 and SQuAD2.0}
    \label{table:2}
    \end{table}
\subsection{Document QA}
\begin{wrapfigure}{r}{0.4\textwidth} %this figure will be at the right
    \includegraphics[width=0.4\textwidth]{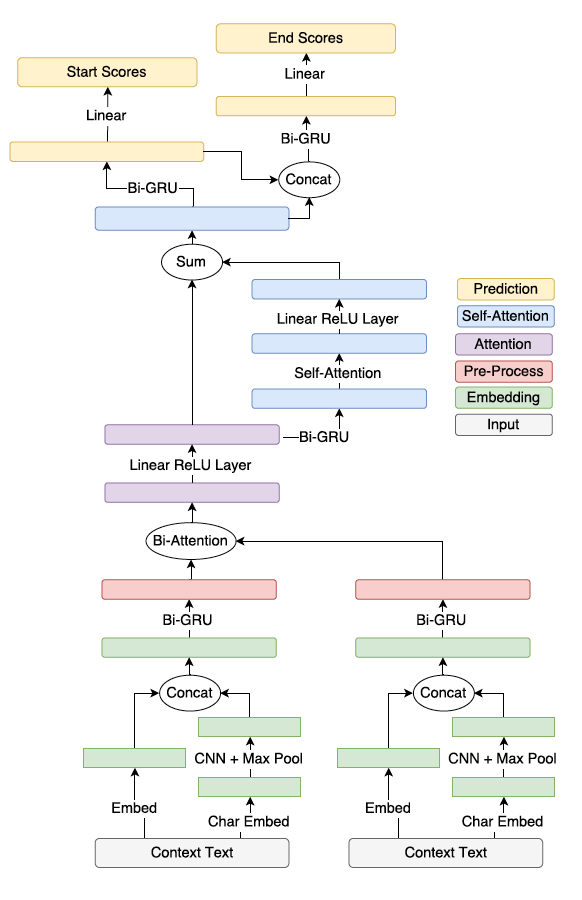}
      \centering
      \caption{DocQA Structure [\citenum{clark_simple_2017}]}
      \label{fig:docqastruc}
\end{wrapfigure}

\subsubsection{Structure of the Document QA model [Fig \ref{fig:docqastruc}]}
First, the model embeds each word into character-level and word-level embeddings and forward the concatenation of those two embeddings into the pre-process layer. Then, a bidirectional GRU [\citenum{cho_learning_2014}] in the pre-processing layer, which is considered as a variation of the LSTM, maps the question and document embeddings to context-aware embeddings. 

\subsubsection{The 4 output vectors from the Attention layer}
A.the vector for the context word $h_i$ B.attended vector $c_i$ for each context token C.element-wise product for context word with its attended vector $h_i\odot c_i$ D.element-wise product for the query-to-context vector with its attended vector $q_c\odot c_i$. \\
$c_i$: Attended vector for each context token [\citenum{clark_simple_2017}]: 
\begin{align*}
p_{ij} = \frac{e^{a_{ij}}}{{\sum_{j=1}^{n_q} e^{a_{ij}}}},\ c_{i} = \sum_{j=1}^{n_q} p_{ij}\cdot q_j
\end{align*}
$q_c$: The query-to context vector [\citenum{clark_simple_2017}]: 
\begin{align*}
m_i = \max_{1 \leq j \leq n_q} a_{ij},\ p_i = \frac{e^{m_{i}}}{{\sum_{i=1}^{n_c} e^{m_{i}}}},\ q_c = \sum_{i=1}^{n_c} h_{i}\cdot p_i
\end{align*}
These outputs will be further analyzed by the following layers to find the exact position of the answer spans.

\subsubsection{Pros and cons for each confidence method}
For the Original Confidence method, it uses the un-normalized and un-exponentiated score given to each span, which is the sum of the start and end score given to its start and end token, as the confidence score. However, this method does not require the training objective to figure out whether the confidence scores are comparable between paragraphs. So, this method may perform poorly. Thus, we tried the Shared-Normalization method. The training objective is modified where the normalization factor in the softmax operation can be shared by all paragraphs from the same document.  To test whether exposing the model to more text can help it to ignore irrelevant text, we propose the Merge method. It concatenated all the paragraphs during sampling procedure, and add a paragraph separator token with learned embedding vector before each paragraph.

\subsubsection{In Table \ref{table:3}(hyper-parameters in Appendix)}
We have based our implementation of the DocQA model on the code released at https://github.com/allenai/document-qa. We mainly tuned hyper-parameters and all results are included in https://drive.google.com/drive/folders/1aSS0fdCiQi0WYOMMweU8VrnEPm40-vbI?usp=sharing. The following table shows the hyper-parameters. Since the input data is in sequence, the exponential weighted moving average(EMA) is used to reduce the noise for the sequence input data. For each model, We pick the set of hyper-parameters that performs the best and include the f1 scores for those in the table in the main report.
\begin{table}[h!]
    \centering
     \begin{tabular}{||c c c c c c||} 
     \hline
      Model & Batch size & Num of epoch & EMA & learning rate & Dropout rate\\ 
     \hline\hline
      None(paragraph-level)&  45 & 26 & 0.999 and 0.9 & 1.0 & 0.8\\ 
      Shared normalization&  45 & 26 & 0.999 & 1.0 & 0.7 and 0.8\\ 
      Merge&  45 & 26 & 0.999 & 1.0 &  0.8\\
      Original confidence&  45 & 26 & 0.999 & 1.0 &  0.8\\
     \hline\hline
    \end{tabular}
    \caption{Table to Doc Experiments}
    \label{table:3}
\end{table}

\subsubsection{Novel idea}
\begin{figure}[h]
    \centering
    \begin{minipage}{0.4\textwidth}
        \centering
        \includegraphics[width=0.9\textwidth]{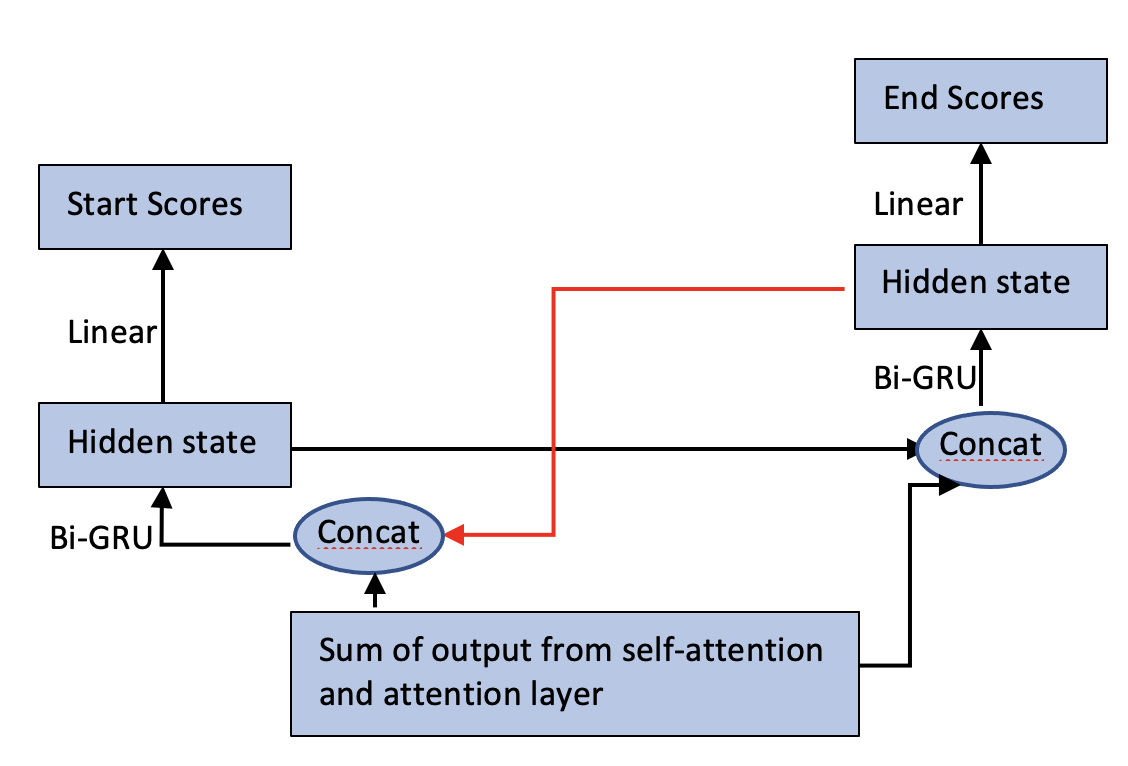}
        \caption{Add connection}
        \label{fig:addconn}
    \end{minipage}\hfill
    \begin{minipage}{0.4\textwidth}
        \centering
        \includegraphics[width=0.9\textwidth]{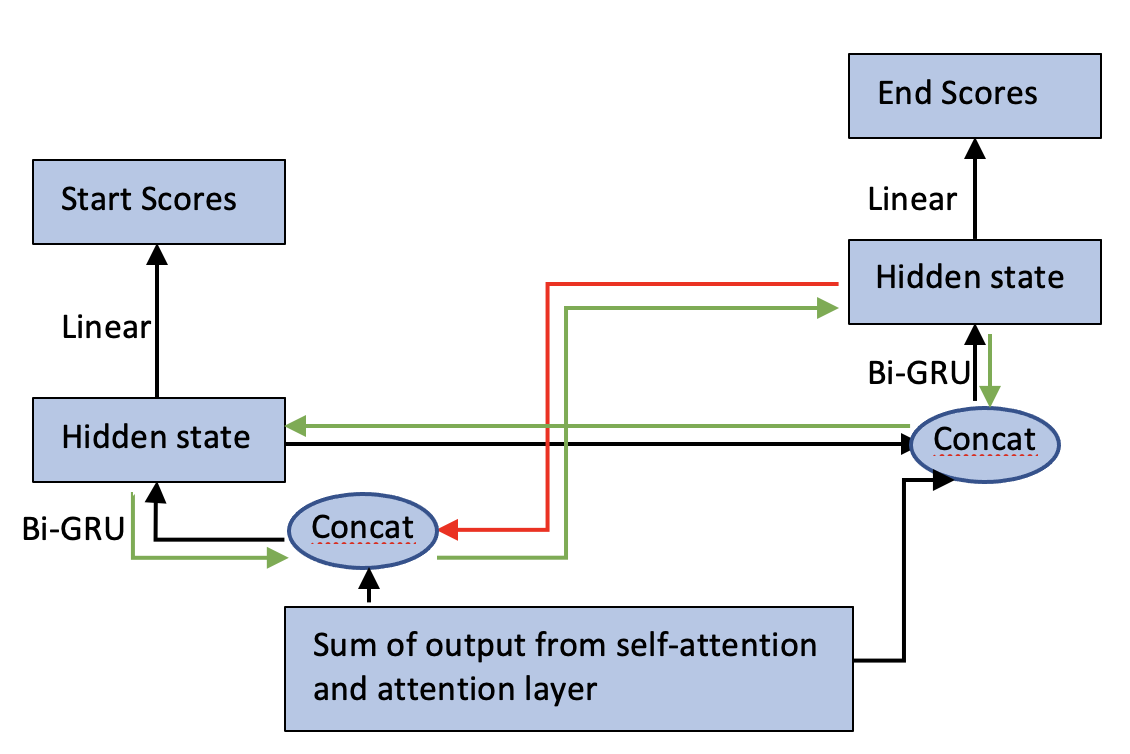}
        \caption{Error signal during back-propagation}
        \label{fig:errorsignal}
    \end{minipage}
\end{figure}

In the left figure (Fig \ref{fig:addconn}), the red arrow shows the connection we added, which is from the hidden state of the answer end token to the input for the bi-GRU for predicting the answer start token. In the right figure (Fig \ref{fig:errorsignal}), the green arrows show the error signal when doing back-propagation.

\subsubsection{Code reference}
The source code is from https://github.com/allenai/document-qa. We tuned the hyper-parameters in the \verb|ablate.squad.py|. The model structure is created in \verb|ablate-triviqa.py|. We also tuned some parameters in the setting of the model. There is a repository in google drive where we include all the results from training models with different modes, tuning hyper-parameters, and the colab codes for creating the plot for the experiments part. Repository: https://drive.google.com/drive/folders/1aSS0fdCiQi0WYOMMweU8VrnEPm40-vbI?usp=sharing

\subsection{ALBERT and the Retro-Reader}

\subsubsection{ALBERT}
Since the announcement of BERT in late 2018 [\citenum{devlin_bert_2019}], researchers have largely relied on increasing model size to improve performance on downstream tasks [\citenum{liu_roberta_2019}, \citenum{yang_xlnet_2020}].  While this strategy works for research labs supported by large cloud vendors, language models that require GPU memory beyond what is general available on a single server have limited real-world applications.  The goal of ALBERT (A Lite BERT) [\citenum{lan_albert_2020}] is to drastically reduce the number of parameters required by BERT, while providing a similar level of performance.

ALBERT uses two methods to reduce the number of parameters -- parameter sharing and factorized embedding.  With respect to task of question-answering, ALBERT only performs slightly worse than BERT (SQuAD2.0 F1 score:  BERT-base 80.4, Albert-base 80.0 [\citenum{lan_albert_2020}]).

\paragraph{Parameter Sharing}
There are 12 attention layers in the base version of BERT, each with its own set of parameters.  In the base version of ALBERT however, all 12 layers share the same set of weights (Fig.\ref{fig:albert01Layers} in Appendix).  As a result, ALBERT has about 10x less parameters than BERT.

\paragraph{Factorized Embedding}
ALBERT ’s second way of achieving model size reduction is through factorized embedding.  By decomposing the large vocabulary embedding matrix into 2 smaller matrices, the size of the vocabulary embeddings is no longer tied to the size of the hidden layer (Fig.\ref{fig:albert02EmbeddingMatrix} in Appendix).  The memory requirement for the embedding matrix is reduced from O(V x H), to O((V x E) + (E x H)).  This can be a significant reduction in memory requirement if H >> E.

\begin{figure}[h]
    \centering
    \begin{minipage}{0.45\textwidth}
        \centering
        \includegraphics[width=0.9\textwidth]{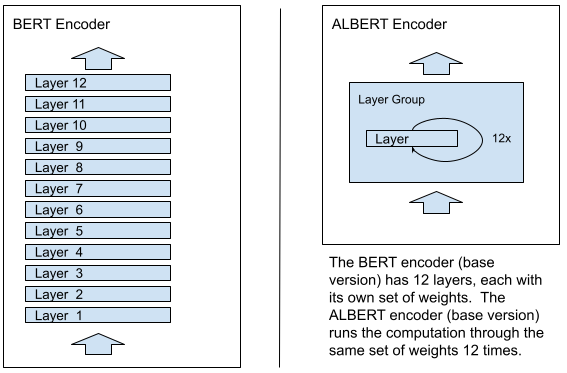}
        \caption{ALBERT parameter sharing.}
        \label{fig:albert01Layers}
    \end{minipage}\hfill
    \begin{minipage}{0.45\textwidth}
        \centering
        \includegraphics[width=0.9\textwidth]{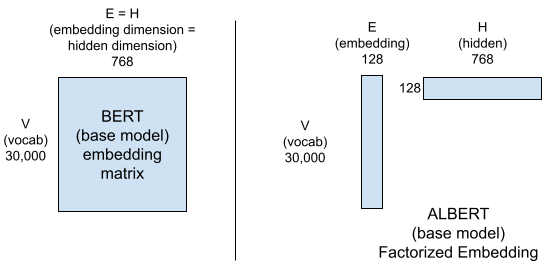}
        \caption{BERT embedding matrix vs ALBERT factorized embedding.}
        \label{fig:albert02EmbeddingMatrix}
    \end{minipage}
\end{figure}

\subsubsection{Retro Reader}
Near the top of the SQuAD2.0 leaderboard [\citenum{noauthor_stanford_nodate}] is Shanghai Jiao Tong University’s Retro-Reader [\citenum{zhang_retrospective_2020}].  The Retro-Reader uses an ensemble of models, one of which is based on ALBERT.  We have re-implemented their non-ensemble ALBERT model.  In order to speed up our research and make the best use of computing resources, we have chosen to use “albert-base-v2” [\citenum{lan_albert_2020}] instead of the much bigger “albert-xxlarge-v2” used by the Jiao Tong team.

\paragraph{Answer or No-Answer}
The prediction head that is fit on top of the ALBERT pre-trained model in order to perform the SQuAD2.0 task has 2 components.  To determine whether a question is answerable, the hidden state of the first output token, which corresponds to [CLS] in the input, is fed through a linear layer with softmax activation.  Cross entropy loss is used to train the model to learn whether a question is answerable.

\paragraph{Answer Span}
The start and end markers of an answer requires our model to predict a span.  The hidden state of all output tokens, of sequence S (e.g. 512) in length, is fed through a linear layer to calculate logits.  The result is a 2 x S matrix which signifies the model’s prediction of which token marks the start and which token marks the end of the answer span.  Cross entropy loss is used.

\begin{figure}[h]
  \includegraphics[scale=0.5]{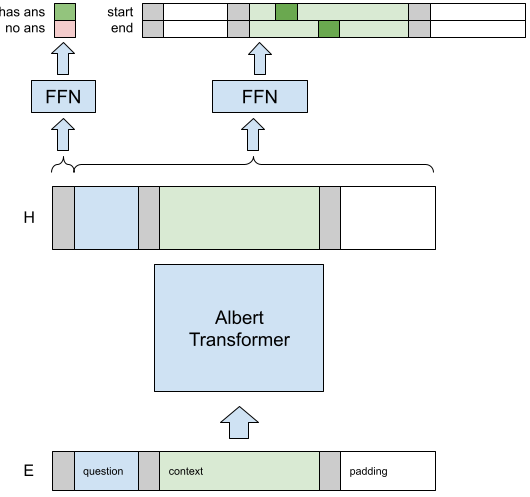}
  \centering
  \caption{Our implementation of the Retro-Reader based on the Albert base model.}
  \label{fig:albert03RetroReader}
\end{figure}

The above losses are then averaged over the batch.  Please refer to Fig \ref{fig:albert03RetroReader} for a depiction of our implementation.

\subsubsection{Hyperparameter Tuning of ALBERT Retro-Reader}

To keep the comparison between BIDAF, DocumentQA and ALBERT meaningful, we performed hyperparameter tuning using the smallest pre-trained ALBERT model available from Hugging Face [\citenum{wolf_huggingfaces_2020}].  We made simplifications from the original Retro-Reader codebase [\citenum{zhang_retrospective_2020}] in order to carry out our experiments.  We performed fine-tuning using the full SQuAD2.0 training dataset, and validated our results using SQuAD2.0’s full dev dataset.  We focused on hyperparameter tuning of batch size, learning rate and sequence length.  We also explored the effect of the re-initialization of weights of the last linear layer.

\paragraph{Batch Size}

\begin{figure}[h]
  \includegraphics[scale=0.5]{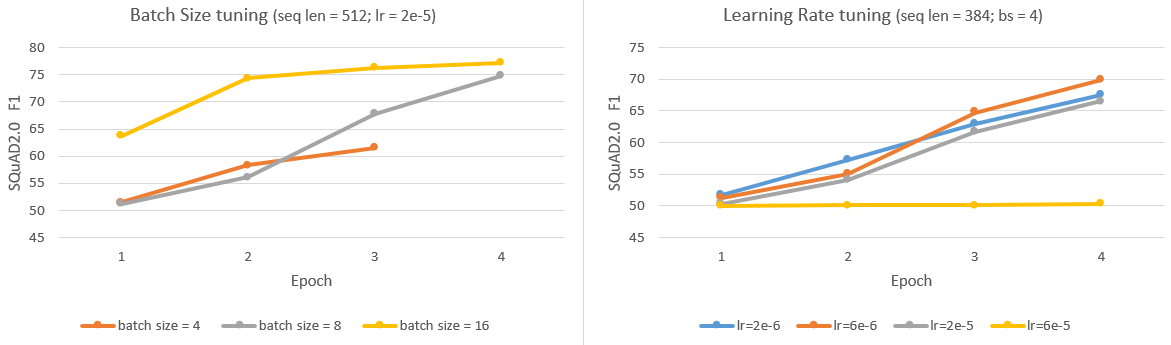}
  \centering
  \caption{ALBERT Retro-Reader: tuning batch size and learning rate}
  \label{fig:albert11batchsizeLearningrate}
\end{figure}

Fig \ref{fig:albert11batchsizeLearningrate} shows the effect of changing batch size on the F1 of the SQuAD2.0 dev test dataset.  We have fixed the input sequence length at 512, with the Adam optimizer’s [\citenum{kingma_adam_2017}] learning rate at 2e-5.  As expected, the larger batch size of 16 provides better F1 score.  We were unable to experiment on batch size of 32 due to memory limitation.

\paragraph{Learning Rate}
The Retro-Reader team [\citenum{zhang_retrospective_2020}] focused their training using a learning rate of 2e-5 for their Adam optimizer.  We have experimented with learning rates around that value, using increments of around 3 times.  Fig \ref{fig:albert11batchsizeLearningrate} shows the F1 performance of learning rates of 2e-6, 6e-6, 2e-5, 6e-5.  We can see that, with our smaller base ALBERT model, a lower learning of 6e-6 actually performs better than the typical 2e-5.

\begin{figure}[h]
  \includegraphics[scale=0.5]{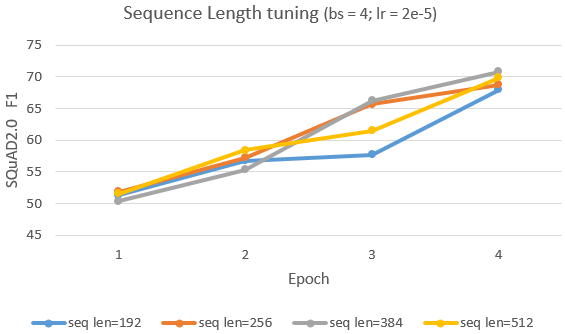}
  \centering
  \caption{ALBERT Retro-Reader: tuning sequence length}
  \label{fig:albert13seqlen}
\end{figure}

\paragraph{Sequence Length}
Fig \ref{fig:albert13seqlen} shows the effect of different input sequence length on SQuAD2.0’s dev set F1 score.  In pre-training an MLM model, longer sequence length usually results in a better performing pre-trained model.  However, when fine-tuning for a given task, the optimal input sequence length typically is a reflection of the average length of the input dataset.  In the case of SQuAD2.0, the average paragraph (“context”) is 118 words and the average question is 11 words in length.  Taking sub-tokenization into account, we can see that extending the sequence length beyond 256, to 384 and 512, does not necessarily improve the F1 score.

\subsubsection{Our implementation of the ALBERT Retro-Reader}
We have based our implementation of the ALBERT Retro-Reader [\citenum{zhang_retrospective_2020}] on the code released by the original authors at the Shanghai Jiao Tong University at https://github.com/cooelf/AwesomeMRC.  Our modifications mainly revolve around drastically simplifying their configuration to only support the ALBERT model.  Instead of almost 200 source files, our implementation only requires 3 python files.

The Jiao Tong University implementation in turn relied heavily on source code from Hugging Face [\citenum{wolf_huggingfaces_2020}] at https://github.com/huggingface/transformers.  For example, the Jiao Tong team has modified the ALBERT implementation by Hugging Face to add the "answer / no answer" prediction head.  The Jiao Tong team also reused evaluation code provided by SQuAD 2.0 [\citenum{rajpurkar_know_2018}].  We have relied on the same validation code for our reported metrics such as F1.

\section{Attributions}
Zhipeng came up with the idea of using SQuAD 2.0 as the dataset of our project.  He is also responsible for the experiments on the BIDAF model.  Ryan is also our resident LaTeX expert.  Tianjiao was the founding member of this team.  She managed the experiments on the Document QA model with different confidence methods, and compare the performance on Squad1.1 between the BIDAF model and Document QA mode. She also tried to figure out whether adding more connections in the prediction layer can better the performance. Marshall extracted the initial smaller datasets from SQuAD 2.0 for prototyping, and drafted common sections of this report.  He managed the experiments on the ALBERT Retro-Reader model, and studied several papers to come up with the novel idea of specific re-initialization of a parameter-shared network to speed up fine-tuning.
%---------------------------------------------------------------------------------

\newpage
\medskip
\bibliography{references}

\end{document}